# Res-CR-Net, a residual network with a novel architecture optimized for the semantic segmentation of microscopy images.

*Hassan Abdallah[6‡], Asiri Liyanaarachchi[4‡], Maranda Saigh[4‡], Samantha Silvers[4‡], Suzan Arslanturk[3,2], Douglas J. Taatjes[7], Lars Larsson[8], Bhanu P. Jena[2,4,5], Domenico L. Gatti[1,2]*

[1]Department of Biochemistry, Microbiology and Immunology, School of Medicine, [2]NanoBioScience Institute, [3]Department of Computer Science, College of Engineering, [4]Department of Physiology, School of Medicine, [5]Center for Molecular Medicine & Genetics, School of Medicine, [6]Department of Mathematics, College of Science, Wayne State University, Detroit, MI 48201, USA; [7]Department of Pathology and Laboratory Medicine, Microscopy Imaging Center, University of Vermont College of Medicine, Burlington, VT 05405, USA; [8]Karolinska Institute, Stockholm, Sweden.

[‡] These authors, listed in alphabetical order, contributed equally to the study.

**Correspondence:** E-mail: dgatti@med.wayne.edu

*Deep Neural Networks (DNN) have been widely used to carry out segmentation tasks in both electron and light microscopy. Most DNNs developed for this purpose are based on some variation of the encoder-decoder type U-Net architecture, in combination with residual blocks to increase ease of training and resilience to gradient degradation. Here we introduce Res-CR-Net, a type of DNN that features residual blocks with either a bundle of separable atrous convolutions with different dilation rates or a convolutional LSTM. The number of filters used in each residual block and the number of blocks are the only hyperparameters that need to be modified in order to optimize the network training for a variety of different microscopy images.*

In recent years the task of semantic segmentation has attracted significant interest in biomedical fields where automated annotation of images is an important process for the extraction of functional information or for 3D reconstruction [1]. Since first introduced, Fully Convolutional Networks (FCN)[2], in which the last fully connected layers are replaced with convolutional layers, have steadily improved the accuracy of semantic segmentation. Additional improvements have been achieved with the introduction of atrous convolutions in DeepLab models [3, 4], and the widespread adoption of U-Net models[5, 6] with encoder-decoder architecture. A conditional random field (CRF) [7] has been added to some deep neural networks (DNN) as a post processing step in order to refine the final segmentation [3, 4], or embedded in the NN itself in the form of 'CRF-as-Recurrent Neural Network (RNN)' [8] for an end-to-end segmentation task [9].

The idea behind the U-Net architecture is that segmentation can be conceptually decomposed in two operations: 1) semantic content extraction in the encoder arm of the NN, and 2) Progressive addition/replacement of the extracted semantic content to the original size image in the decoder arm of the NN. However, it is intuitively hard to understand why these two operations cannot proceed smoothly and progressively in a pixel-wise fashion, without the need for first shrinking and then re-expanding the concept field.

Here, we introduce a type of residual NN [10], Res-CR-Net, that combines residual blocks based on separable atrous *C*onvolutions (the *C* in *CR*), with a final residual block based on *R*NNs [11] (the *R* in *CR*), instead of a CRF, to refine the segmentation task by exploiting the smooth transition between adjacent rows and columns in images. This network displayed competitive performance with respect to a traditional Res-U-Net, when tasked to segment images from either electron or light microscopy in three separate categories, using for training only a small number of images with 15-fold augmentation.

*Datasets*
1. Electron microscopy (EM) gray scale images (262 x 400) of rat liver cells (Fig. 1a), obtained as described in [12]. The segmentation task was to recognize nuclei (red mask in Fig. 1b), mitochondria (blue mask in Fig. 1b), and anything else that is neither nuclei or mitochondria (green mask in Fig. 1b). In this case, Res-CR-Net was trained for 90 epochs with 8 images/ground truth masks (corresponding to 1 batch), with 15-fold augmentation in each epoch, and validated against 4 images/ground truth masks.

2. Fluorescence microscopy (FM) rgb images (300 x 300) of human skeletal muscle cells from biopsies stained using myosin heavy chain antibody conjugated to a fluorescent fluorophore (Fig. 1c) [LL and BPJ, manuscript in preparation]. The segmentation task was to recognize the regions of myofibers cytoplasm (red mask in Fig. 1d), the boundaries between myofibers (green mask in Fig. 1d), and gaps between the myofibers or empty spaces (blue mask in Fig. 1d). In this case, Res-CR-Net was trained for 90 epochs with 10 images/ground truth masks (corresponding to 1 batch), with 15-fold augmentation in each epoch, and validated against 6 images/ground truth masks.

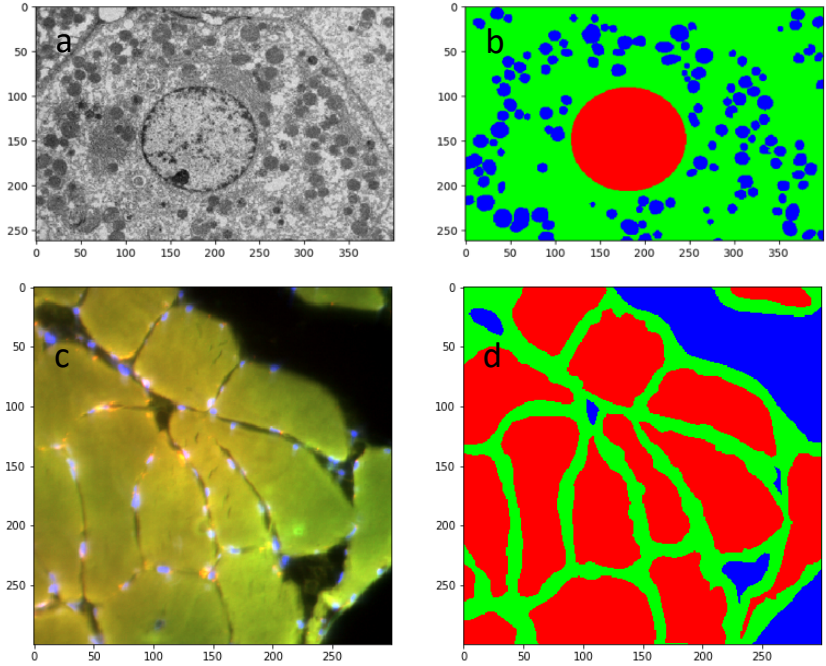

**Fig. 1. a.** Typical electron micrograph of rat hepatocytes, and **b.** Ground truth mask (nuclei, red; cytoplasm, green; mitochondria, blue). **c.** Typical fluorescence micrograph of a human skeletal muscle biopsy section stained for myosin heavy chain, and **d.** Ground truth mask (myofiber cytoplasm, red; boundaries, green; gaps/empty spaces, blue).

*Data augmentation*

To avoid overfitting and to decrease the time and labor involved in hand labeling the regions of interest in the dataset images, we relied on geometric data augmentation. Each pair of image and ground truth mask was sheered or rotated at random angles, shifted with a random center, vertically or horizontally mirrored, and randomly scaled in/out. The parts of the image left vacant after the transformation were filled in with reflecting padding. During training, each epoch called on 15 steps (batches), and a batch consisted of all images in the dataset (thus achieving 15-fold augmentation/epoch). We found this training strategy to be more effective, although slightly slower, than the alternative strategy of using smaller batches, with each epoch processing the entire dataset, and training for a larger number of epochs.

*Architecture of Res-CR-Net.*

A flowchart of the architecture of Res-CR-Net is shown in Fig. 2. It combines two types of residual blocks to boost the performance of per-pixel classification, while maximizing modularity and thus flexibility in adapting the network to different segmentation tasks.

1. CONV RES BLOCK. The traditional U-Net backbone architecture, with its the encoder-decoder paradigm, is replaced by a series of modified residual blocks, each consisting of three parallel branches of separable + atrous convolutions with different dilation rates, that produce feature maps with the same spatial dimensions as the original image. As stressed in [4], the rationale for using multiple-scale layers is to extract object features at various receptive field scales. Res-CR-Net offers the option of concatenating or adding the parallel branches inside the residual block before adding them to the shortcut connection. In our test, concatenation produced the best result. A Spatial Dropout layer follows each residual block. A slightly modified STEM block processes the initial input to the network. *n* CONV RES BLOCKS can be concatenated.

2. LSTM RES BLOCK. A new type of residual block features a residual path with two orthogonal bidirectional 2D convolutional Long Short Term Memory (LSTM) layers (Fig. 2).

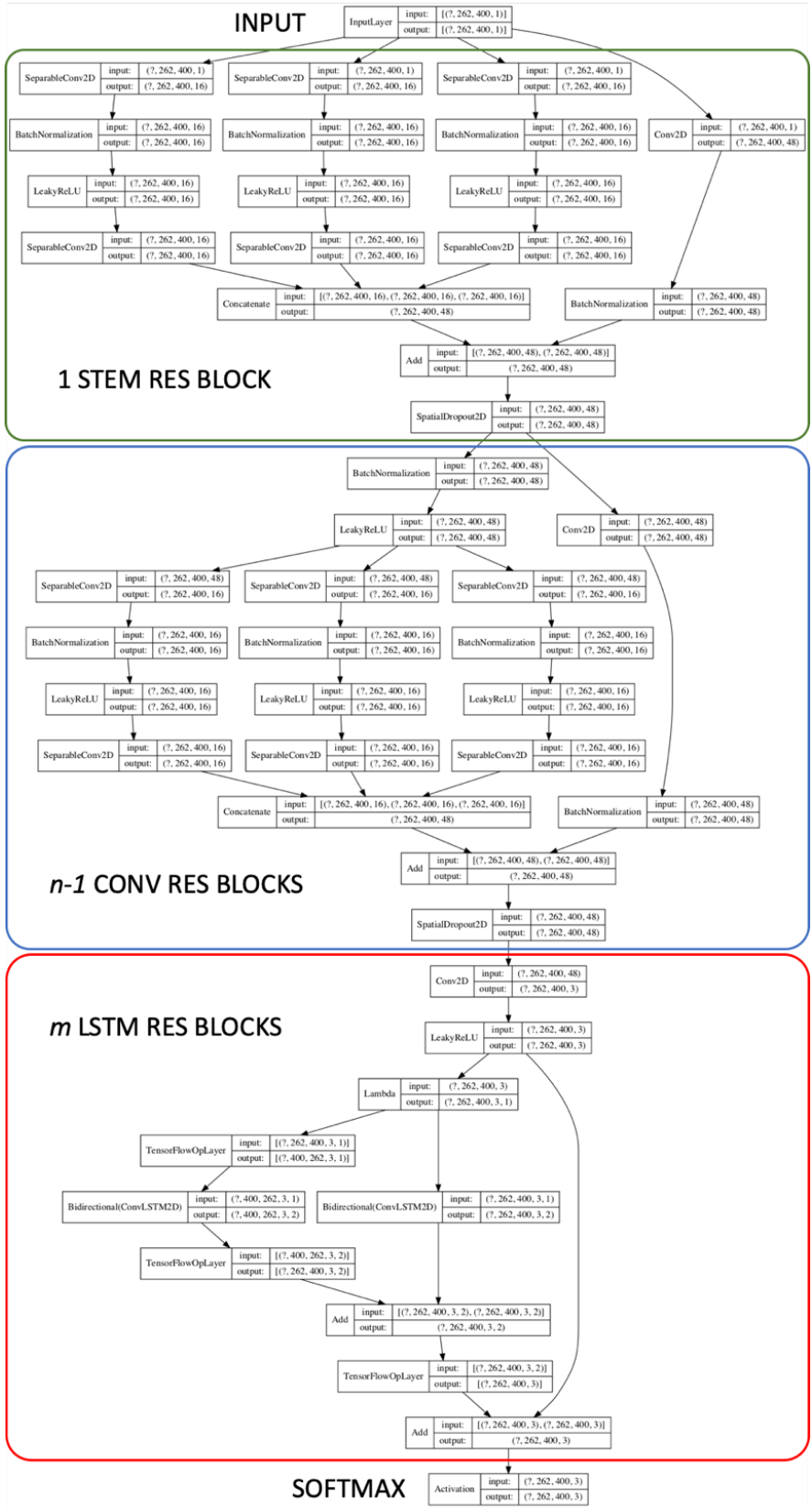

**Fig. 2.** Architecture of Res-CR-Net. In this example the net processes images of dimensions [262,400,3], there are *n* = 6 CONV RES BLOCKS, and *m* = 1 LSTM RES BLOCK.

For this purpose, the feature map 4D tensor emerging from the previous layer first undergoes a virtual dimension expansion to 5D tensor (i.e. from [4,260,400,3] [batch size, rows, columns, number of classes] to [4,260,400,3,1]). In the example of Fig. 2 the 2D LSTM layer treats 260 consecutive tensor slices of dimensions [400,3,1] as the input data at each iteration. Each slice is convolved with a single filter of kernel size [3,3] with 'same' padding, and returns a slice of the exact same dimension. In one-direction mode the LSTM layer returns a tensor of dimensions [4,260,400,3,1]. In bidirectional mode it returns a tensor of dimensions [4,260,400,3,2]. The intuition behind using a convolutional LSTM layer for this operation lies in the fact that adjacent image rows share most features, and image objects often contain some level of symmetry that can be properly memorized in the LSTM unit. This type of intuition is perhaps somewhat related to the idea of representing the mean-field iterations of a CRF as a Recurrent Neural Network [9]. Since the same intuition applies also to image columns, the expanded feature map of dimensions [4,260,400,3,1] from the earlier part of the network is transposed in the 2$^{nd}$ and 3$^{rd}$ dimension to a tensor of dimensions [4,400,260,3,1]. In this case the LSTM layer processes 400 consecutive tensor slices of dimensions 260,3,1 as the input data at each iteration, returning a tensor of dimensions [4,400,260,3,2] which is transposed again to [4,400,260,3,2]. The two LSTM output tensors are then added and the final dimension is collapsed by summing its elements, leading to a final tensor of dimensions [4,260,400,3] which is added to the shortcut path. *m* LSTM RES BLOCKS can be concatenated.

A *LeakyReLU* activation is used throughout Res-CR-Net. After the last residual block a *softmax* activation layer is used to project the feature map into the desired segmentation. Res-CR-Net uses as part of the loss function a variation of the *Dice coefficient*, $D(\hat{y}, y)$ [13-17]:

$$D(\hat{y}, y) = \frac{2 \sum_i \hat{y}_i y_i + s}{\sum_i \hat{y}_i + \sum_i y_i + s}$$

known as the *Tanimoto coefficient with complement* [18], $\tilde{T}(\hat{y}, y)$:

$$\tilde{T}(\hat{y}, y) = \frac{T(\hat{y}, y) + T(1 - \hat{y}, 1 - y)}{2}$$

in which $T(\hat{y}, y)$ is:

$$T(\hat{y}, y) = \frac{\sum_i \hat{y}_i y_i + s}{\sum_i (\hat{y}_i^2 + y_i^2) - \sum_i \hat{y}_i y_i + s}$$

$\hat{y} \equiv \{\hat{y}_i\}$, $\hat{y}_i \in [0,1]$ are the probabilities for the *i*-th pixel, $y \equiv \{y_i\}$, $y_i \in \{0,1\}$ are the corresponding ground truth labels, and *s* is a smoothing scalar. Accordingly, the *Tanimoto loss* is defined as:

$$L_{\tilde{T}}(\hat{y}, y) = 1 - \tilde{T}(\hat{y}, y)$$

which can be weighted appropriately to account for class imbalance. In Res-CR-Net weights are derived with a "contour" aware scheme, by replacing a step-shaped cutoff at the edges of the mask foreground with a border that tends to separate touching objects of the same class [6].

Res-CR-Net was implemented using *Keras* [19, 20] deep learning library running on top of *TensorFlow* 2.1 [21]. Training and testing tasks were conducted on the Wayne State University HPC grid equipped with Intel i7 CPUs and Tesla V100-SXM2-16GB GPUs. Res-CR-Net can process images of any size, without constrains imposed by the down-pooling and up-sampling operations of an encoder-decoder architecture. Furthermore, thanks to its modular design (both type of blocks can be repeated as needed), the depth of the network can be changed at will by modifying only the number of residual blocks, and the number of filters, kernel sizes, and dilation rates in each residual block.

*Results.*

When trained for 90 epochs on the EM dataset of 8 gray scale images with 15-fold augmentation (see above), Res-CR-Net (configured with 6 atrous residual blocks and 1 LSTM residual block) achieved ~90% segmentation accuracy (*Tanimoto coefficient*: 91.6%) on the 4 images of the validation set (Fig. 3).

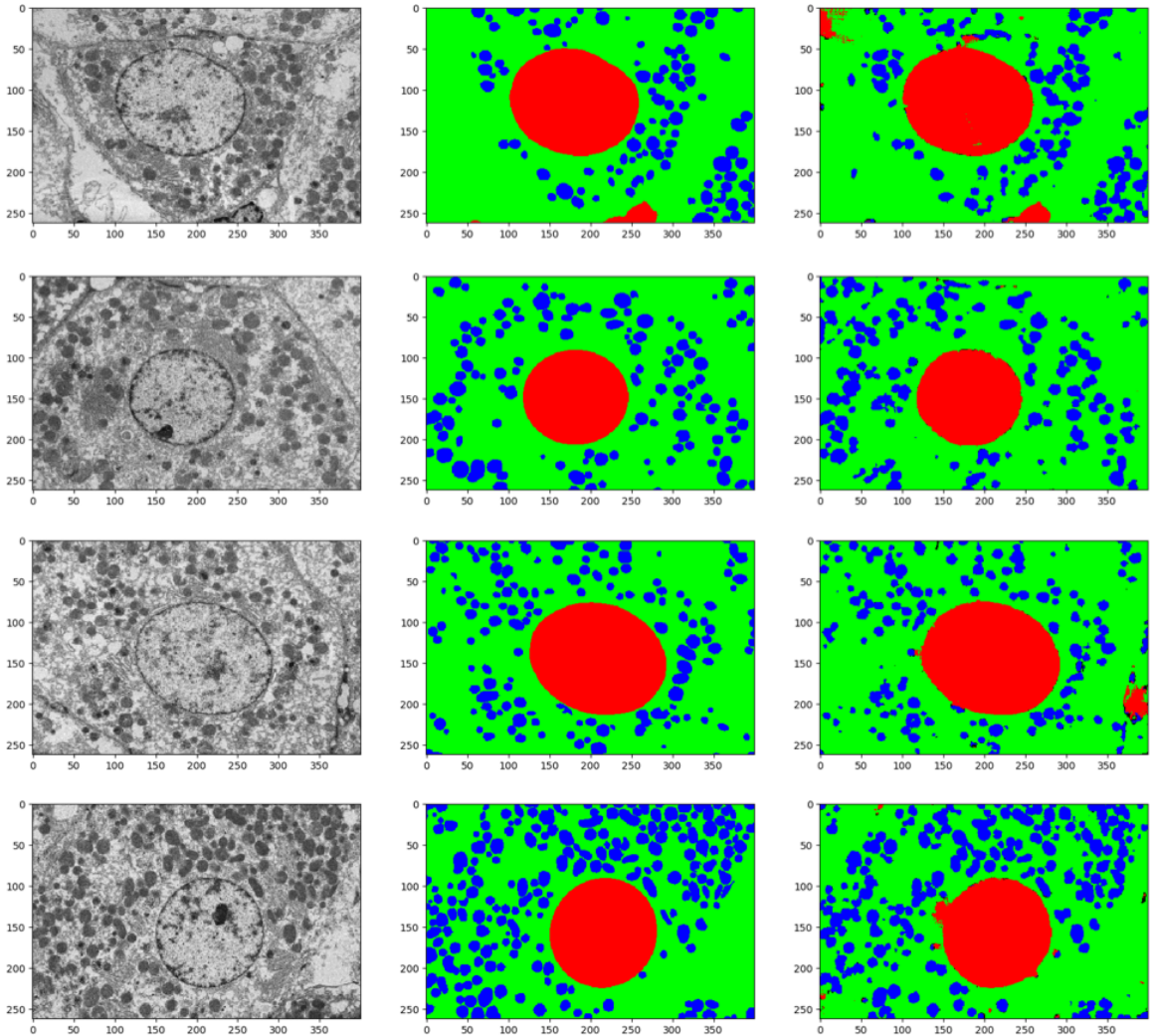

**Fig. 3.** EM validation set. From left to right, electron micrographs of rat liver cells, ground truth masks (nuclei, red; cytoplasm, green; mitochondria, blue), and Res-CR-Net predicted masks.

The segmentation task in these images is to identify both the cell nucleus and the mitochondria. It's worth noting that segmentation of mitochondria in electron micrographs from focused ion beam scanning electron microscopy (FIB-SEM) or automated tape-collecting ultramicrotome scanning electron microscopy (ATUM-SEM) has been the focus of multiple reports [22-26] in recent years. However, due to the variety of mitochondrial structures, as well as the presence of noise, artifacts and other sub-cellular structures, mitochondria segmentation in EM has proven to be a difficult and challenging task. As apparent from Fig. 3, Res-CR-Net performs very well in this task, despite the small number of images used for training.

When trained for 90 epochs on the FM dataset of 10 images with 15-fold augmentation (see above), Res-CR-Net (identically configured as for the EM dataset) also achieved ~90% segmentation accuracy (*Tanimoto coefficient*: 90.7%) on the 6 images of the validation set (Fig. 4):

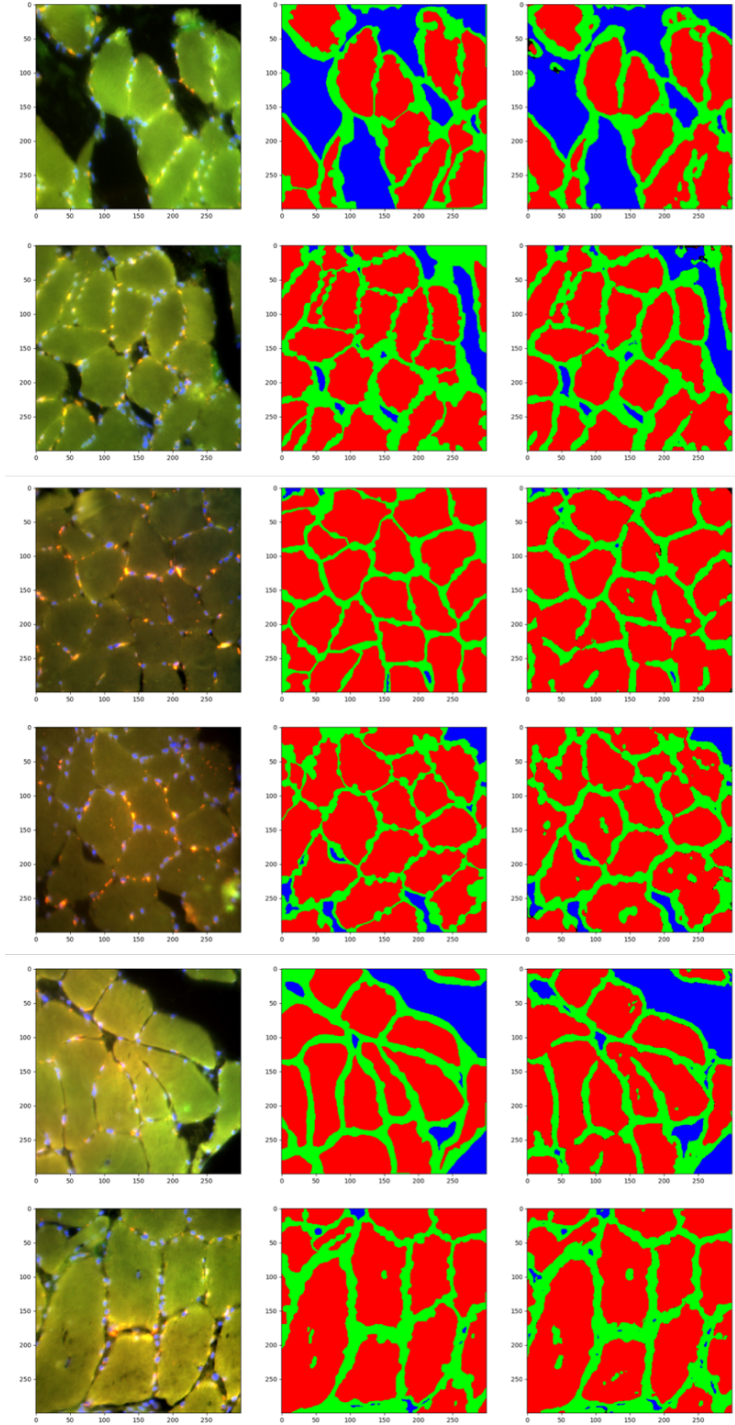

**Fig. 4.** FM validation set. From left to right, fluorescence micrographs of human skeletal biopsy sections stained for myosin heavy chain, ground truth masks (cytoplasm, red; myofibers boundaries, green; gaps/empty spaces, blue), Res-CR-Net predicted masks.

In this task Res-CR-Net showed excellent refinement properties as attested by its convergence speed (~40 epochs) and stability (up to 90 epochs) with respect to both the training and validation set (Fig. 5).

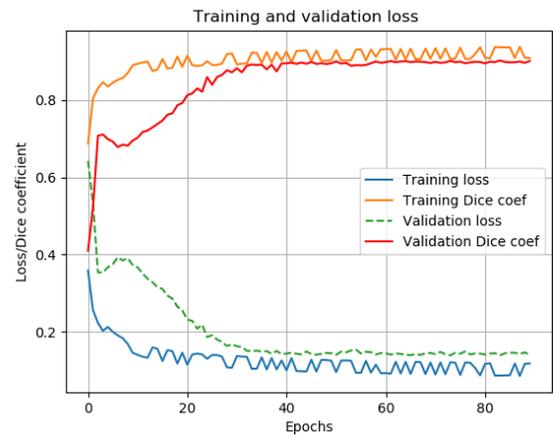

**Fig. 5.** Training and validation Loss and Dice coefficient versus epochs for Res-CR-Net processing of FM images of human muscle cell from biopsies.

Other metrics about both segmentation tasks, defined as:

$Jaccard\ index = \frac{TP}{(TP+FP+FN)}$

$Precision = \frac{TP}{(TP+FP)}, \quad Recall = \frac{TP}{(TP+FN)},$

$F1\ score = \frac{2*precision*recall}{precision+recall}$

are shown in Table 1:

| *Table 1.* | *Dice coeff.* | *Jaccard ind.* | *Precision* | *Recall* | *F1 score* |
|---|---|---|---|---|---|
| *EM* val. set | 0.899 | 0.859 | 0.926 | 0.922 | 0.924 |
| *FM* val. set | 0.888 | 0.844 | 0.917 | 0.914 | 0.915 |

*Conclusion.*

In this report we present Res-CR-Net, a neural network featuring a novel FCN architecture, with very good performance in multiclass segmentation tasks of both electron (gray scale, 1 channel) and light microscopy (rgb color, 3 channels) images of relevance in the analysis of pathology specimens. The network was effective in achieving segmentation of the validation set images almost indistinguishable from the ground truth, as attested by the very high values (~90%) of the Tanimoto coefficients. In particular, Res-CR-Net offers some advantages with respect to other networks inspired to an encoder-decoder architecture, as it is completely modular, with residual blocks that can be proliferated as needed, and it can process images of any size and shape without changing layers size and operations. Res-CR-Net can be particularly effective in segmentation tasks of biological images, where the labeling of ground truth classes is laborious, and thus the number of annotated/labeled images in the training set is small.

**Author Contributions:** HA,SA,DLG developed the architecture of Res-CR-Net. MS,SS produced all ground truth masks. AL,DJT,LL,BPJ provided the image datasets. HA,LL,SA,BPJ,DLG wrote the manuscript. All authors participated in discussions and proofreading the manuscript.

**Acknowledgements:** This study was supported in part by the National Science Foundation grants EB00303, CBET1066661 (BPJ), and the Erling-Persson Family Foundation, the Swedish Research Council grant 8651, and the Stockholm City Council grants Alf 20150423 and 20170133 (LL).

**Software:** Source code for Res-CR-Net is deposited at: https://github.com/dgattiwsu/Res-CR-Net

**Corresponding Author**
E-mail: dgatti@med.wayne.edu

**ORCID**
Domenico Gatti: 0000-0002-6357-3530
Suzan Arslanturk: 0000-0002-4554-4373
Bhanu Jena: 0000-0002-6030-8766
Lars Larsson: 0000-0003-3722-035x
Hassan Abdallah: 0000-0003-1256-0446
10